\documentclass[letterpaper, 10 pt, conference]{ieeeconf}
\IEEEoverridecommandlockouts
\usepackage{amsmath,amsfonts,amssymb}
\usepackage{algpseudocode}
\usepackage{algorithm}
\usepackage{array}
\usepackage{svg}
\usepackage[caption=false,font=normalsize,labelfont=sf,textfont=sf]{subfig}
\usepackage{textcomp}
\usepackage{stfloats}
\usepackage{url}
\usepackage{verbatim}
\usepackage{graphicx}
\usepackage{cite}
\usepackage{wrapfig}
\usepackage{multirow}
\usepackage{microtype}
\usepackage{tabularx,ragged2e}
\usepackage{booktabs}
\usepackage[letterpaper, left=0.667in, right=0.667in, bottom=0.597in, top=0.792in]{geometry}
\usepackage{fancyhdr}

\makeatletter
\let\NAT@parse\undefined
\makeatother
\usepackage{hyperref}

\begin{document}

\title{\LARGE \bf 
Learning Vehicle Dynamics from Cropped Image Patches for Robot Navigation in Unpaved Outdoor Terrains
}

\author{Jeong Hyun Lee, Jinhyeok Choi, Simo Ryu, Hyunsik Oh, Suyoung Choi, and Jemin Hwangbo$^*$%
\thanks{This work was supported by the Agency For Defense Development Grant Funded by the Korean Government (UD210024ID)}%
\thanks{The authors are with Robotics and Artificial Intelligence Lab in the Department of Mechanical Engineering, KAIST, Daejeon 34141, Republic of Korea %
{\tt\small {\{joshualee, first2021, simoryu, ohsik1008, swimchoy, jhwangbo\}}@kaist.ac.kr}
}
}

\maketitle
\thispagestyle{empty}
\pagestyle{empty}

\thispagestyle{fancy}
\fancyhf{}
\renewcommand{\headrulewidth}{0pt}
\cfoot{This work has been submitted to the IEEE for possible publication. Copyright may be transferred without notice, after which this version may no longer be accessible.}

\begin{abstract}
In the realm of autonomous mobile robots, safe navigation through unpaved outdoor environments remains a challenging task.
Due to the high-dimensional nature of sensor data, extracting relevant information becomes a complex problem, which hinders adequate perception and path planning.
Previous works have shown promising performances in extracting global features from full-sized images. However, they often face challenges in capturing essential local information.
In this paper, we propose Crop-LSTM, which iteratively takes cropped image patches around the current robot's position and predicts the future position, orientation, and bumpiness.
Our method performs local feature extraction by paying attention to corresponding image patches along the predicted robot trajectory in the 2D image plane. This enables more accurate predictions of the robot's future trajectory.
With our wheeled mobile robot platform \textit{Raicart}, we demonstrated the effectiveness of Crop-LSTM 
for point-goal navigation in an unpaved outdoor environment. 
Our method enabled safe and robust navigation using RGBD images in challenging unpaved outdoor terrains.
The summary video is available at https://youtu.be/iIGNZ8ignk0.
\end{abstract}


\section{Introduction}

In recent years, autonomous mobile robots have been studied for tasks such as transportation \cite{fraedrich2019transportation}, surveillance \cite{weisser1999autonomous}, and search and rescue \cite{yurtsever2020survey}. 
In such missions, the robots often need to navigate through challenging and rugged terrains, requiring them to determine safe pathways.
To this end, many researchers \cite{dolgov2008practical}, \cite{chu2012local} leveraged model-based path planning algorithms to identify suitable paths to avoid potential dangers.
For instance, Hu et al. \cite{hu2018dynamic} utilized dynamics equations to calculate the future position of the vehicle, finding an optimal path to avoid collision with obstacles.

However, model-based approaches sometimes suffer in autonomous navigation on uneven terrain due to the complex nonlinear vehicle dynamics involved.
To express the nonlinear vehicle dynamics, model-based approaches \cite{ostafew2014nonlinear_dynamics_mpc}, \cite{williams2016mppi} utilized mathematical models to represent the system behavior.
However, modeling the system proves challenging due to the terrain's deformable nature, the compressibility of tires, and the propensity of the tires to slip.
These complexities can hinder navigation on complex and unpredictable terrains.

In response to these complexities, researchers introduced learning-based methods, which can acquire vehicle dynamics data from extensive driving experience and eliminate the necessity of complex formulas associated with model-based approaches. 
One of the popular methods in this approach is learning from experts \cite{bojarski2016end-to-end}, \cite{kebria2019imitation} where the agent imitates the expert's driving demonstration. 
This method enables the robot to learn from experienced drivers and leverage their ability to navigate complex environments.
The expert's ability to leverage the semantic meanings of the complex environment can aid the robot in making informed decisions. 
However, it requires an extensive amount of expert-driving data, which is challenging and expensive to collect.

\begin{figure}[!t]
\centering
\includegraphics[width=\linewidth]{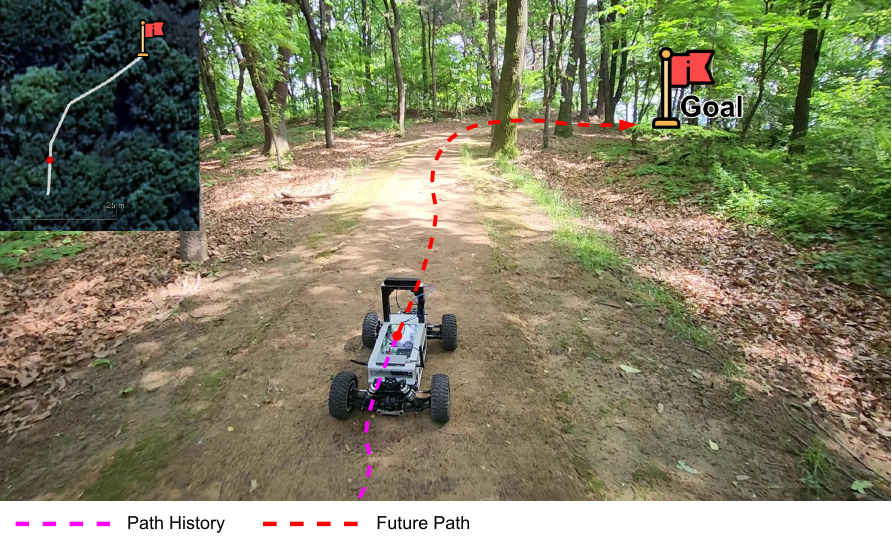}
\caption{The wheeled robot platform, \textit{Raicart}, adeptly navigates through unpaved outdoor terrain. 
The environment is a straw blanket on a dirt road. 
\textit{Raicart} navigates this challenging course without explicit object segmentation. 
The proposed navigation system finds a path that minimizes bumpiness and oscillation in orientation.}
\vspace{-0.25cm}
\label{fig_0}
\end{figure}

The utilization of Supervised Learning (SL) can address the limitations inherent in Imitation Learning (IL) approaches.
Some works\cite{kahn2018gcg}, \cite{wellhausen2019should} utilized SL and used diverse and randomly generated data sets, eliminating the need to accumulate extensive expert-driving data required in IL. 
For instance, Khan et al. \cite{kahn2021badgr} proposed a method utilizing SL to learn the vehicle dynamics model from randomly controlled driving data, effectively simplifying the data collection process.

Although SL is widely utilized for navigation tasks, extracting the necessary information from high-dimensional exteroceptive sensor data remains a challenge.
Sensor abstraction is crucial for SL in autonomous navigation because it reduces the complexity of raw sensor data, enabling effective feature extraction and learning of relevant patterns.
To extract significant details of the surrounding environments, some researchers  \cite{kendall2019learn_in_a_day}, \cite{shah2021ving}, \cite{shah2021recon}, \cite{shah2022viking} employed Convolutional Neural Networks (CNN) to encode camera images into small vectors and utilized the encoded data for the navigation task.

However, all methods mentioned above focused on extracting global information rather than utilizing attention mechanisms. 
By incorporating attention mechanisms, researchers have shown the capability of extracting significant local information.
For instance, Guan et al. \cite{guan2018crop_diagnose} utilized self-attention to generate masks and identify critical regions within the image. 
Similar approaches were conducted on image classification \cite{dosovitskiy2020vit}, \cite{wang2017residualattention} and object detection tasks \cite{perreault2020spotnet} and showed outstanding performances.

Drawing inspiration from these methods, we propose a cropping technique that pays attention to the regions along the predicted robot trajectory in the 2D image plane. 
Through the application of the cropping method, we trained an accurate vehicle dynamics model.
Subsequently, we developed a navigation system that reduced the bumpiness experienced during the drive, enabling the robot to navigate through unpaved outdoor terrains safely.

Our main contributions are as follows:
\begin{itemize}
\item{We propose Crop-LSTM, a novel architecture that utilizes cropping at the feature map. The introduction of cropping allows the development of a more precise vehicle dynamics model compared to previous approaches that solely utilize full-sized images.}
\item{We present a navigation system featuring a trained Crop-LSTM. 
The sampler generates multiple sequences of future throttling velocity and steering angle commands.
From this, the robot can select and execute the sequence that minimizes bumpiness and oscillation in orientation.}
\item{We built a wheeled robot platform \textit{Raicart} upon a 1/5 scale off-road RC car chassis.
The robot includes an NVIDIA Jeston AGX Xavier computer with proprioceptive and exteroceptive sensors. 
We share the details of the robot in this paper.}
\item{We conducted autonomous navigation experiments on unpaved outdoor environments with an in-house wheeled robot platform, \textit{Raicart}.}
\end{itemize}

\section{Related work}
Autonomous navigation presents challenges due to the highly nonlinear dynamics of vehicles \cite{ostafew2014nonlinear_dynamics_mpc}. 
The intricate interactions of the wheel and the terrain can lead to complex and unpredictable behaviors, making precise control difficult.
Researchers \cite{guan2022ga}, \cite{weerakoon2022terp} have approached this by assuming a simplified dynamics model, transforming the navigation problem into a path planning problem. 
For instance, Dolgov et al. \cite{dolgov2008practical} employed a cost map to identify the optimal path through a path planning algorithm. 
Moreover, Chu et al. \cite{chu2012local} utilized a sampling-based optimization technique, generating candidate paths and selecting one that minimizes the user-defined cost.
However, these approaches have limitations, as the robot may struggle to track the planned path consistently where the complex and unpredictable environment leads to deviations from the desired trajectory.

To overcome these limitations, deep-learning techniques were introduced and demonstrated effectiveness in enhancing the robot's ability to navigate challenging terrains.
For instance, Cai et al. \cite{cai2020drift-driving} utilized a Reinforcement-Learning (RL) based controller to replace complex motion equations for aggressive drift driving.
Likewise, Weerakoon et al. \cite{weerakoon2022terp} trained a neural network that generates an attention mask applied to the elevation map, allowing the identification of feasible navigation trajectories. 
Harnessing human-derived data presents an additional avenue for the implementation of deep-learning techniques in robot navigation.
This approach has yielded impressive driving results by utilizing CNNs to extract information from images and incorporating expert driving data \cite{bojarski2016end-to-end}, \cite{kebria2019imitation}.

Semantic segmentation aids driving by precisely classifying distinct regions within an image \cite{romera2017erfnet}, \cite{wang2022sfnet}.
This capability allows mobile robots to identify dangerous regions, enabling them to determine safe paths for navigation.
For instance, Guan et al. \cite{guan2022ga} presented an effective method for semantic segmentation in unpaved outdoor terrains.
They utilized semantic segmentation to classify the navigability levels of the terrain.
However, applying semantic segmentation to mobile robots presents challenges.
Terrains often have varying and unpredictable features, making it difficult for semantic segmentation models to classify distinct regions accurately.

To address these concerns, researchers proposed an alternative end-to-end navigation approach that eliminates the need for explicit image region segmentation.
For instance, Bojarski et al. \cite{bojarski2016end-to-end} proposed a method that directly generates control action without additional sampling or optimization by imitating human driving data.
Additionally, Kenda et al. \cite{kendall2019learn_in_a_day} proposed an end-to-end driving method that leverages simulation and RL instead of human demonstrations.
However, these approaches are limited to navigation on relatively simple and short road scenarios.

To overcome these problems, an alternative method leveraging learned dynamics models to build a navigation system was proposed.
For instance, Khan et al. \cite{kahn2021badgr} introduced a learning-based navigation system that utilizes encoded image data to learn the dynamics model that predicts the robot's future position and evaluates the likelihood of potential collisions and bumpiness.
Building upon this pioneering work, researchers have delved deeper into point-goal navigation by applying topological maps \cite{shah2021ving} and learned latent variable models \cite{shah2021recon}. 
Shah et al. \cite{shah2022viking} showcased the navigational capabilities at kilometer-scale distances.
Moreover, Shah et al.\cite{shah2023lm-nav} harnessed pre-trained large language models to enhance their navigation performance. 

\section{Method}
In the following sections, we provide detailed information about the four major parts of our work. 
A. The wheeled robot platform, \textit{Raicart}, specifically designed for autonomous navigation in unpaved outdoor terrain. 
B. Data collection and auto-labeling process, achieved without human supervision.
C. Crop-LSTM, our novel architecture for efficient learning of the vehicle dynamics model.
D. Our navigation system for safe robot navigation on unpaved outdoor terrain.

\subsection{Raicart: The Wheeled Mobile Robot Platform}

\begin{figure}[!ht]
\centering
\includegraphics[width=\linewidth]{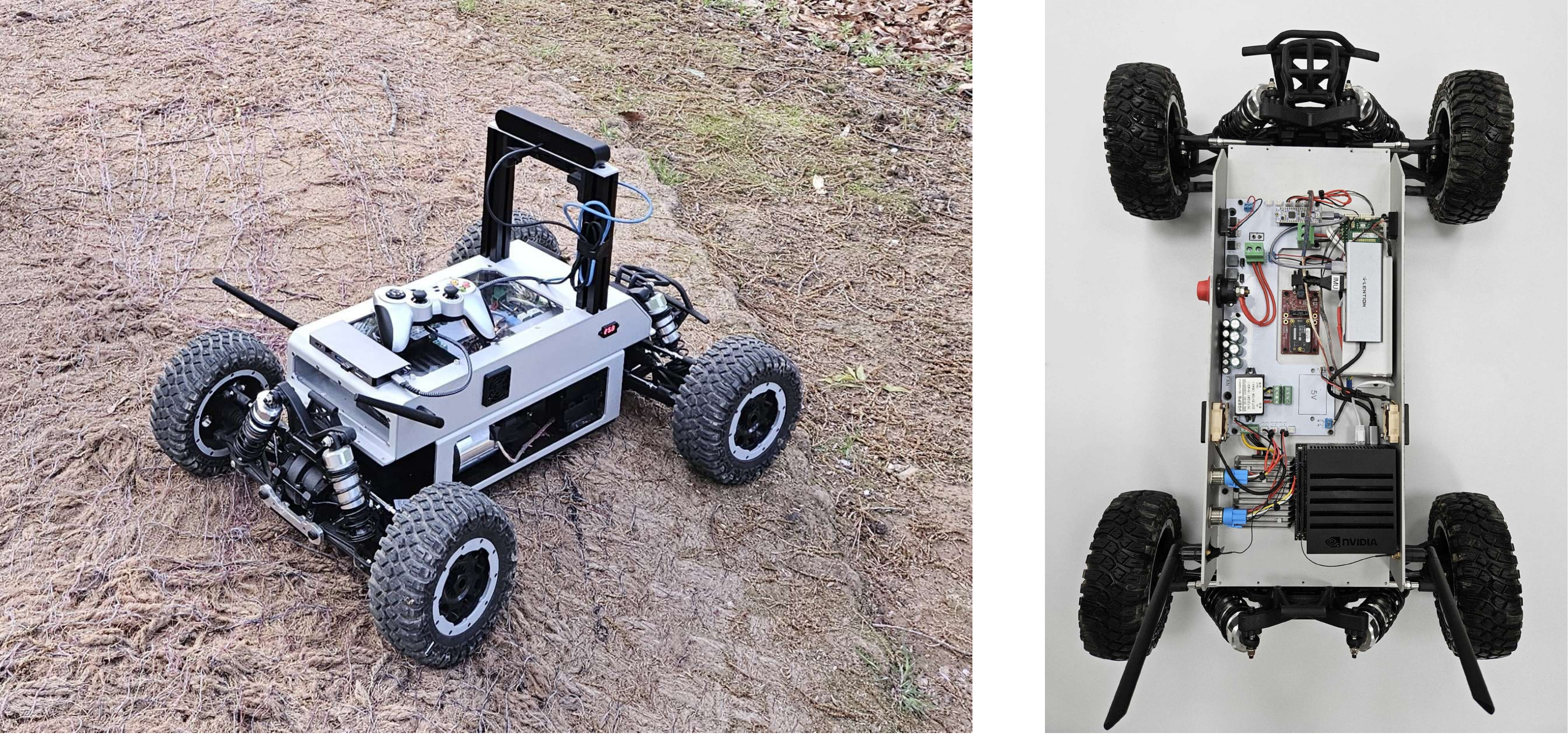}
\caption{The wheeled mobile robot platform \textit{Raicart}}
\vspace{-0.25cm}
\label{fig_raicart}
\end{figure}

\textit{Raicart} is our in-house wheeled robot platform designed for outdoor navigation tasks, shown in Fig. \ref{fig_raicart}. 
The dimensions of the robot are 770mm x 500mm x 480mm, and the weight is 14kg.
The chassis of the robot is constructed over a Losi 1/5 scale RC car, drawing similarities to prior work \cite{goldfain2019autorally}, \cite{o2020f1tenth}.
The system adeptly changes its heading direction through Ackerman steering, effectively preventing the tires from slipping sideways while maneuvering through curves.
The robot is equipped with NVIDIA Jetson AGX Xavier, a small but powerful embedded PC.
We employed the Microstrain 3DM-CV5-IMU for measuring linear acceleration and angular velocity. 
The Zed2 stereo camera was utilized for streaming RGB and depth images with a resolution of 336$\times$188. 
Additionally, we employed the Intel realsense T265 tracking camera for Visual-Inertial Odometry (VIO).

\begin{figure*}[!t]
\centering
\includegraphics[width=\linewidth]{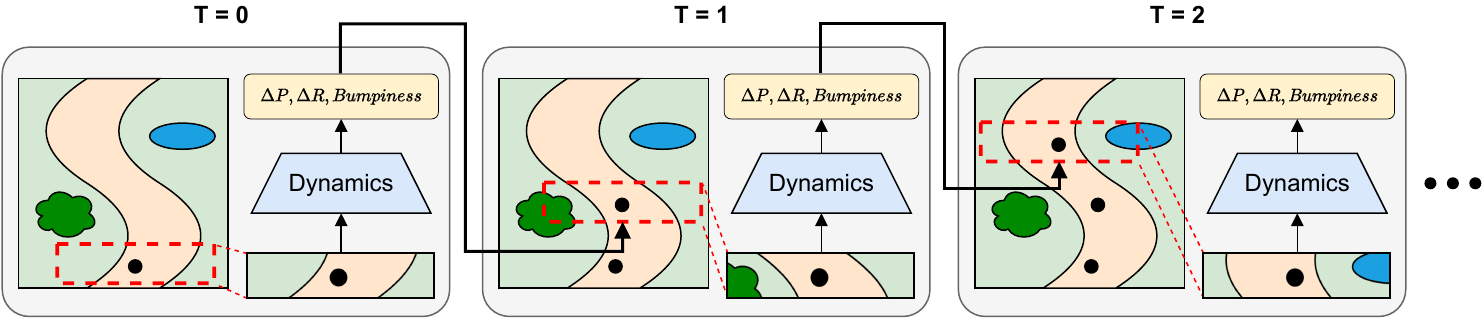}
\caption{Overview of the proposed cropping method. We find the current position on the image and crop the region near the position, shown with a red dotted box. We pass the cropped image patch with a control command to the dynamics model, predicting the next position. We find the new cropping region, and the whole process is repeated.}
\vspace{-0.25cm}
\label{fig_cropping_method}
\end{figure*}

\subsection{Data Collection and Automated Labeling}

For vehicle dynamics model learning, we collected data by driving \textit{Raicart} on unpaved outdoor terrain. 
During the operation, a human intentionally provided random control commands of throttling velocity and steering angle to the robot at a frequency of 10Hz.
All sensor data was collected at the same frequency, totaling 128 episodes of 100,000 data points.

We divided the collected data into event sequences spanning a time horizon of 3 seconds with a timestep of 0.3 seconds.
Using these sequences, we created labels for the prediction of three future events: position, orientation, and bumpiness.
We calculated the one-step position and orientation transition for the position and orientation labels relative to the previous state.

For the bumpiness label, we differ from the previous work \cite{kahn2021badgr}, which defined bumpiness as \textit{true} when the angular velocity or linear acceleration surpasses a specified threshold.
This approach was acceptable as they focused on urban or flat off-road terrains, with minimal likelihood of surpassing specified limits.
However, our study targets unpaved outdoor terrain that experiences various levels of bumpiness when driving on a straw blanket or getting off the road.
Therefore, we introduce a continuous value for the bumpiness label by redefining it as the variance of z-directional linear acceleration in the recent 1-second period. 
This modification transforms the discrete labels into continuous ones, enabling the system to capture and utilize finer-grained information about the terrain's bumpiness level.

\subsection{Crop-LSTM: Learning the Vehicle Dynamics Model}

\begin{figure*}[!t]
\centering
\includegraphics[width=\textwidth]{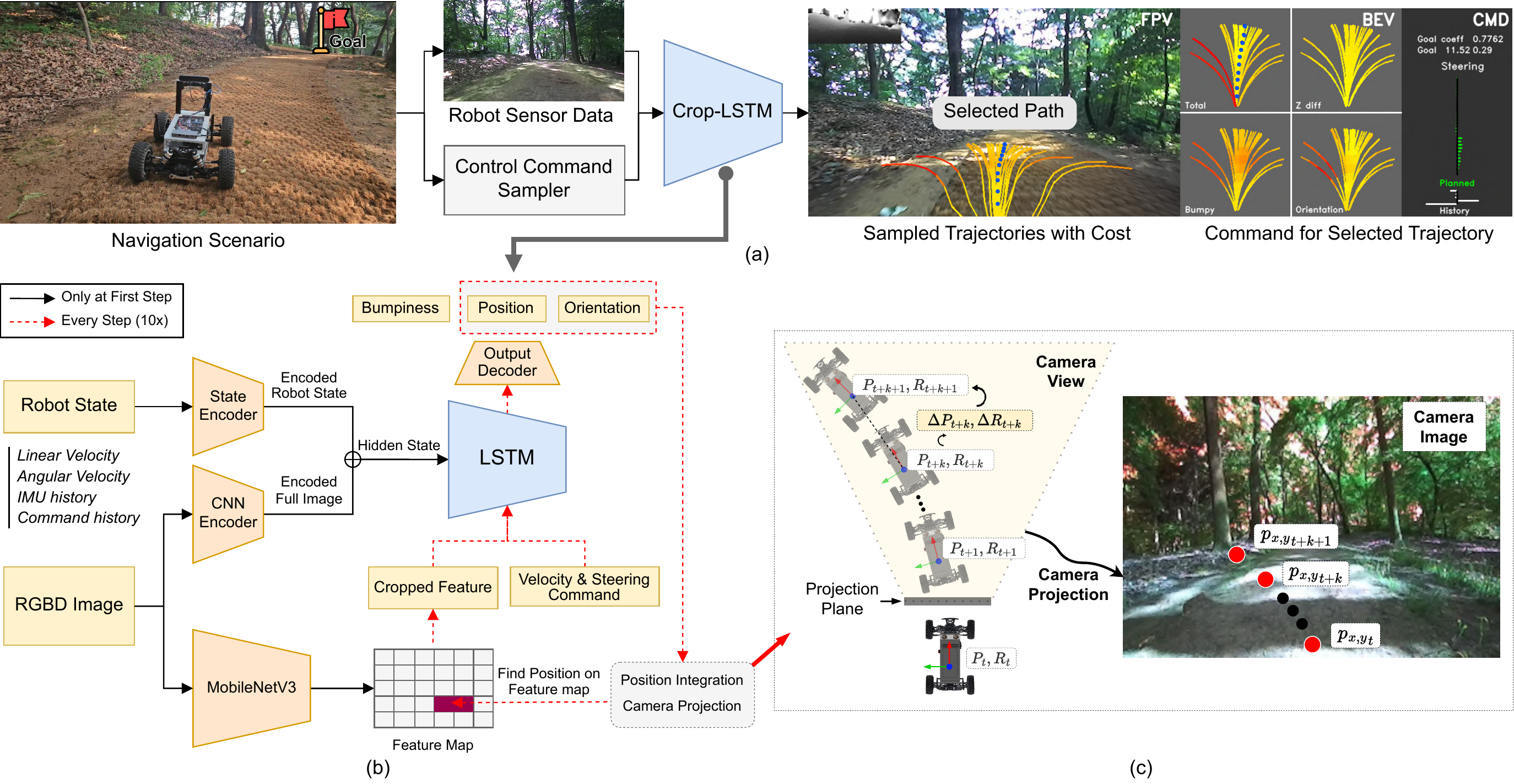}
\caption{(a) The overview of our navigation system. We first collect the robot sensor data.
A sampler generates a batch of candidate control commands.
Subsequently, our proposed vehicle dynamics model, Crop-LSTM, predicts the future position and bumpiness of each sequence of control inputs. 
We find a path that minimizes bumpiness and oscillation in orientation. 
(b) The network architecture of Crop-LSTM. 
In the first step, a full RGB-D image from the Zed2 sensor and robot states are each encoded and concatenated to make an initial hidden state of the LSTM. 
We encode the image to a feature map with a fine-tuned MobileNetV3 encoder (i.e., without the classifier). 
We integrate the previous position and orientation output into the next position. 
The corresponding position in the world frame is projected to the image plane. 
After projection, we crop the features on the feature map near the projected point and pass it to the LSTM encoder. 
(c) Visualization for position integration and camera projection. 
During every step of the inference of the Crop-LSTM, we integrate the predicted state transition into the next position.}
\label{fig_overall}
\end{figure*}

Substantial disparities between predicted and actual vehicle dynamics could result in unexpected maneuvers, underscoring the importance of accurate vehicle dynamics models for mobile robot navigation.
We train an accurate vehicle dynamics model by utilizing the attention mechanism through the novel concept of cropping, as depicted in Fig. \ref{fig_cropping_method}.
By providing explicit attention to the relevant regions through cropping, we enhance the prediction accuracy of the vehicle dynamics model.
Based on this concept, we propose a neural network architecture called Crop-LSTM, illustrated in Fig. \ref{fig_overall}.
For inference of the Crop-LSTM, the initial position is determined as the bottom center of the image.
We then crop the surrounding region from the image and use the dynamics model to predict the subsequent position conditioned on the cropped image patch and control command. 
This process is repeated to predict future events for each step on the horizon.  

We train the Crop-LSTM $f_\theta$ using the collected data set $\mathcal{D}$ to predict the sequence of events $e_{t:t+H}=(e_t, e_{t+1}, ..., e_{t+H})$. 
The event at time $t$ consists of the position $e^0_t$, orientation $e^1_t$, and bumpiness $e^2_t$. 
The network takes in robot states $s_t$ and action sequence $a_{t:t+H}$ as input.
During inference, the predicted future position in the global frame is projected onto the camera frame.
Subsequently, we crop the corresponding region on the feature map generated by the MobileNetV3 encoder \cite{howard2019mobilenetv3} to obtain the cropped feature $c_t$.

We formulate the training objective as minimizing the loss function defined as
\begin{equation}
\label{loss_equation}
\mathcal{L}(\theta, \mathcal{D})=
\sum_{\mathcal{D}}\sum_{k=0}^{2}\alpha_k \cdot \mathcal{L}_{MSE}(f_\theta^k(s_t, a_{t:t+H}), e_{t:t+H}^{k}).
\end{equation}
Using the collected data set, we trained a neural network $f_\theta$ using the Algorithm \ref{alg:train_crop}. The training took about 15 hours on an AMD Ryzen9 5950X processor with an NVIDIA GeForce RTX 4080 Graphics Processing Unit (GPU). 

\begin{algorithm}[H]
\caption{Training Crop-LSTM}\label{alg:train_crop}
\begin{algorithmic}[1]
\Function{Crop }{$P$}
\State project $P_{x,y,z}$ to $p_{x, y}$ in the camera frame
\State find the corresponding position on the feature map
\State select the features c
\EndFunction
\State sample ($s_t, a_{t:t+H}, e_{t:t+H}$) $\in \mathcal{D}$
\State initialize $P_t$, $R_t$ = (0, 0, 0)
\For{{$h \in \{1,\dots,H\}$}}
    \State $c_{t+h} = Crop\,(P_{t+h-1})$ \Comment{Cropping}
    \State $\Delta{P_{t+h}}= f_\theta^0\,(s_t, a_{t+h}, c_{t+h})$ \Comment{Inference}
    \State $\Delta{R_{t+h}}= f_\theta^1\,(s_t, a_{t+h}, c_{t+h})$
    \State $B_{t+h}= f_\theta^2\,(s_t, a_{t+h}, c_{t+h})$
    \State $P_{t+h} = P_{t+h-1} + R_{t+h-1}^T \cdot \Delta{P_{t+h}}$ \Comment{Integration}
    \State $R_{t+h} = R_{t+h-1} \cdot \Delta{R_{t+h}}$
\EndFor
\State calculate loss function using Eqn. (\ref{loss_equation}) and update $\theta$
\end{algorithmic}
\end{algorithm}

\subsection{Navigation System} \label{navigation_system}

Our proposed navigation system is composed of two key components: the learned vehicle dynamics model $f_\theta$ and a cost function $J$.
In the navigation scenario illustrated in Fig. \ref{fig_overall}, the sampler generates a batch of control command sequences. 
Next, we propagate these sampled control commands through the learned vehicle dynamics model $f_\theta$ to obtain candidate trajectories, which encompass predicted position, orientation, and bumpiness at each point along the path.
We select the trajectory that minimizes the cost function $J$, solving the planning problem represented as
\begin{equation}
\begin{aligned}
\label{cost_equation}
& a_{t:t+H}^* = \underset{a_{t:t+H}}{\arg \min} \sum_{point \in Traj}J\,(point) \\
& Traj = (\hat{e}_{t:t+H}^{k}) \subseteq f_\theta^k(s_t, a_{t:t+H}).
\end{aligned}
\end{equation}

Our cost function is designed to ensure safe navigation on unpaved outdoor terrains.
The cost function $J$ is a summation of four sub-costs, $J_{goal}$, $J_{bumpy}$, $J_{ori}$, and $J_z$. 
Each sub-costs serve a specific purpose: guiding the robot toward the goal while simultaneously avoiding excessive bumpiness, sudden changes in orientation, and drastic drops in the z-direction. 
We formulate each sub-costs as
\begin{equation}
\begin{aligned}
\label{cost_definition}
& J_{goal} = \left\| P_{goal} - \hat{e}_{pos} \right\| / \left\| P_{goal} \right\| \\
& J_{bumpy} = \hat{e}_{bumpy} \\
& J_{ori} = \left| \hat{e}_{roll} \right| + \left| \hat{e}_{pitch} \right| + \left| \hat{e}_{yaw} \right| \\
& J_{z} = \left| \hat{e}_{pos_z} \right|.
\end{aligned}
\end{equation}

Utilizing the learned vehicle dynamics model and the cost function, the navigation system operates using the Algorithm \ref{alg:deploy_crop}. 
The onboard NVIDIA Jetson computer performs all computations, and the control command sequence is selected at a frequency of 2Hz, including 0.3 seconds for the inference of the Crop-LSTM. 
To follow the selected throttling velocity and steering angle commands, low-level control for the BLDC motor and the steering servo motor is conducted at a frequency of 10Hz.

\begin{algorithm}[H]
\caption{Deploying Crop-LSTM}\label{alg:deploy_crop}
\begin{algorithmic}[1]
\State \textbf{Input}: Crop-LSTM $f_\theta$, Cost $J$, Goal position
\While{not arrived at goal}
\State update goal position in the current robot frame
\State sample $a_{t:t+H}$
\State get $\Delta{P_{t:t+H}}, \Delta{R_{t:t+H}}, B_{t:t+H} = f_\theta\,(s_t, a_{t:t+H})$
\State solve Eqn. \ref{cost_equation} using Eqn. \ref{cost_definition}
\State execute $a_{t:t+H}^*$
\EndWhile
\end{algorithmic}
\end{algorithm}

\begin{table*}[!t]
\centering
\caption{Training loss results, Mean Squared Error}
\label{table:training_results}
{\tabcolsep=0.3pt\def\arraystretch{1.0}
\begin{tabularx}{\textwidth}{c *9{>{\Centering}p{42pt}}}
\toprule
  & \multicolumn{3}{c}{Training Set}   & \multicolumn{3}{c}{Validation Set} & \multicolumn{3}{c}{Test Set} 
  \tabularnewline \cmidrule(lr){2-4}\cmidrule(lr){5-7}\cmidrule(lr){8-10}
  \textbf{Model} & Position & Orientation & Bumpy & Position & Orientation & Bumpy & Position & Orientation & Bumpy \tabularnewline 
\midrule
  BADGR-Original  & 8.36E-4 &  1.93E-4 &  7.32E-3 & 8.76E-4 & 2.05E-4 & 1.26E-2 & 7.11E-4 &  1.79E-4 & 1.10E-2 \tabularnewline
  BADGR-Modified  &  8.36E-4 & 1.95E-4 &  4.09E-3 &  8.69E-4 & 2.06E-4 & 1.25E-2  & 7.33E-4 &  1.75E-4 &  1.14E-2 \tabularnewline
  Cropping on the image domain  &  5.27E-4 & 7.10E-5 & 3.82E-3 &  6.07E-4 &  9.79E-5 & 7.66E-3  & 7.31E-4 &  \textbf{1.40E-4} &  1.33E-2 \tabularnewline
  Cropping on the feature map (Ours)  &  \textbf{4.81E-4} & \textbf{6.33E-5} & \textbf{2.73E-3} &  \textbf{5.61E-4} &  \textbf{8.79E-5} & \textbf{6.14E-3}  & \textbf{7.05E-4} &  1.49E-4 &  \textbf{1.07E-2} \tabularnewline
\bottomrule
\end{tabularx}}
\end{table*}

\section{Results}
\subsection{Training results}

To quantitatively evaluate the prediction accuracy of the proposed method, Crop-LSTM, we conducted a comparative analysis with the baseline BADGR \cite{kahn2021badgr}, which also trained the vehicle dynamics model.
The specific settings for the experiments are as follows:

\begin{itemize}
\item{BADGR-Original: As a baseline, the model descriptions on the paper \cite{kahn2021badgr} were implemented. A single LSTM encoder and a multi-head MLP were employed for each event prediction.}
\item{BADGR-Modified: We modified the original BADGR model by incorporating two LSTM encoders. This separates the prediction of bumpiness from the prediction of position and orientation.}
\item{Cropping on the image domain: As an ablation study, we perform cropping on the image domain rather than the feature map. At each step of the LSTM encoder, we encoded the cropped image using the MobilenetV3 encoder.}
\item{Cropping on the feature map: This is the model we propose. We employed two LSTM encoders and provided cropped features at each step of the LSTM encoder.}
\end{itemize}

To evaluate the performance of these models, we divided the entire data set into three subsets: training, validation, and test sets.
Among the data set, one trajectory was reserved exclusively for the test set, and the rest of the data set was split in an 8:2 ratio to create the training and validation set.
The models were trained using the training data set, and their prediction performance was compared based on the mean squared error loss across the three subsets.
Each training was conducted three times in a consistent training environment, and the average loss is reported in Table \ref{table:training_results}. 

\begin{figure}[!t]
\centering
\includegraphics[width=\linewidth]{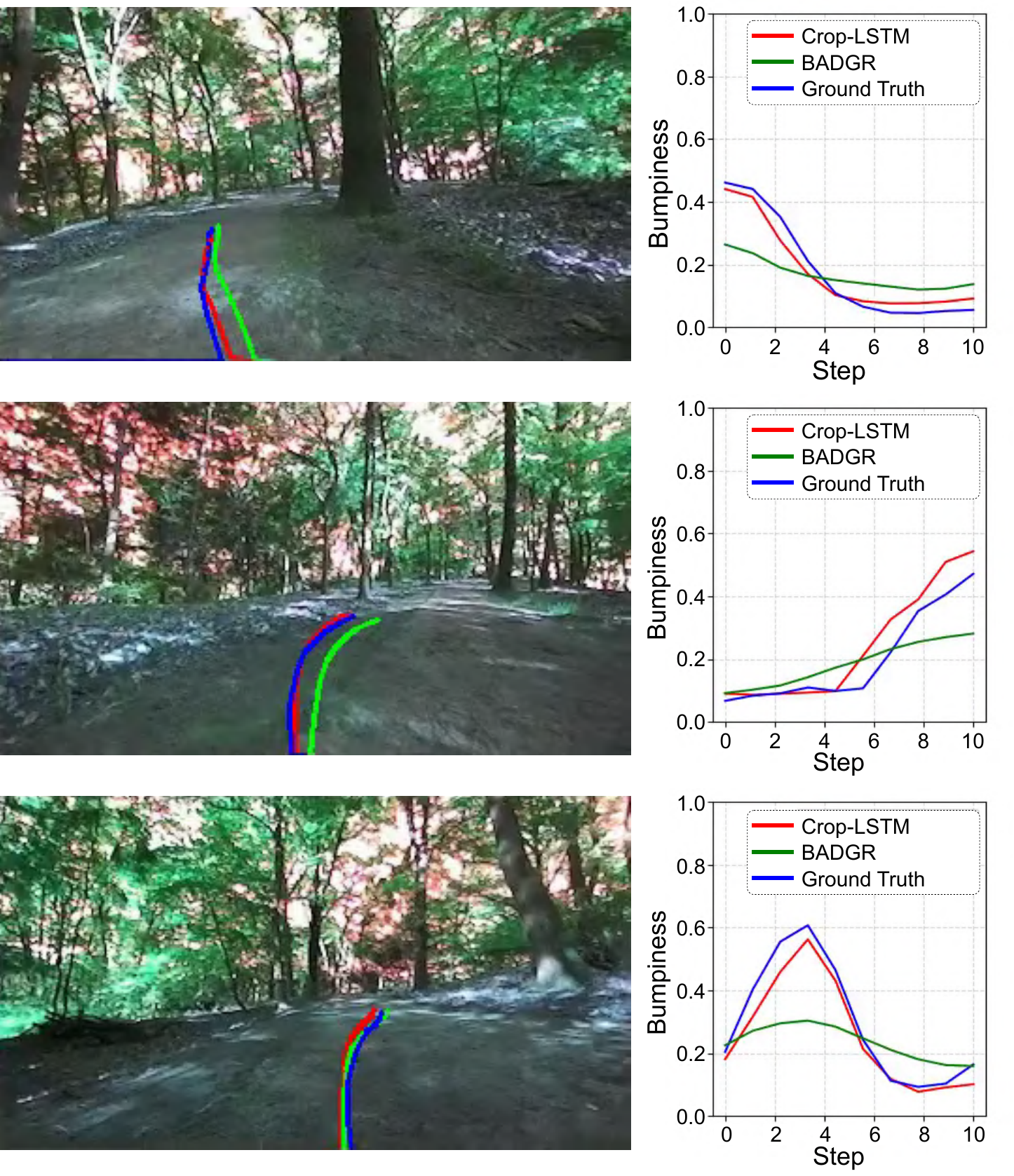}
\caption{Visualized results for the trained Crop-LSTM and BADGR-Modified. The ground-truth data is blue, the prediction of Crop-LSTM is red, prediction of BADGR-Modified is green. The figures are randomly selected results from the data set. Projected positions are on the left, followed by bumpiness on the right. We can find that Crop-LSTM shows accurate predictions on both position and bumpiness.}
\label{fig_visualized_training}
\end{figure}

Our proposed method, involving cropping on the feature map, exhibits the best prediction performance for the mean squared error (MSE) in terms of position, orientation, and bumpiness across the training set and the validation set.
In the test set, our method achieves the second-best results for orientation error, with the top-performing method also employing cropping.
This result shows that cropping is an effective method for increasing the prediction accuracy of the vehicle dynamics model.

Cropping on the image domain also outperforms the baseline, but it requires inference of the Mobilenet encoder for the cropped image patches at every step of the LSTM, whereas cropping on the feature map encodes the image only once at the first step of the LSTM.
This results in an increase in the inference time, making it 1.745 times slower than cropping on the feature map. 
The increase in the inference time can hinder navigation performance as it requires more time to adapt to changes in the surrounding environment.

In Fig. \ref{fig_visualized_training}, we visualize prediction results about the future positions and bumpinesses obtained from the trained Crop-LSTM and BADGR models. 
The output of Crop-LSTM is depicted in red, BADGR's output in green, and the ground truth data in blue. 
While Crop-LSTM demonstrates greater accuracy overall, both Crop-LSTM and BADGR correctly predict the trajectory's tendency. 
The most significant difference lies in the bumpiness prediction. 
Crop-LSTM consistently exhibits lower prediction error, whereas BADGR shows limitations in capturing the fine details.

\subsection{Navigation}
Utilizing our trained Crop-LSTM and cost function, we conducted navigation experiments using Algorithm \ref{alg:deploy_crop} explained in section \ref{navigation_system}. 
The experiment took place two months after the data collection, resulting in varying visual features such as grass amount on the road edge, sunlight levels, and shadow directions. 
These shifts in the testing environments caused the image to differ from the training set, requiring generalization of the Crop-LSTM on unseen environments. 
Despite these challenges, our robot successfully navigated through challenging unpaved outdoor terrain.

\subsubsection{Prediction performance} 
In the testing environment, we assessed the prediction performance of our Crop-LSTM.
We sampled a batch of control command trajectories, propagated them through the Crop-LSTM, and evaluated each trajectory with the cost function. 
This evaluation enabled us to determine the trajectories that effectively accomplish four key objectives: reaching the goal position, avoiding bumpiness, preventing sudden changes in orientation, and avoiding sudden drops in the z-direction.

\begin{figure}[!ht]
\centering
\includegraphics[width=\linewidth]{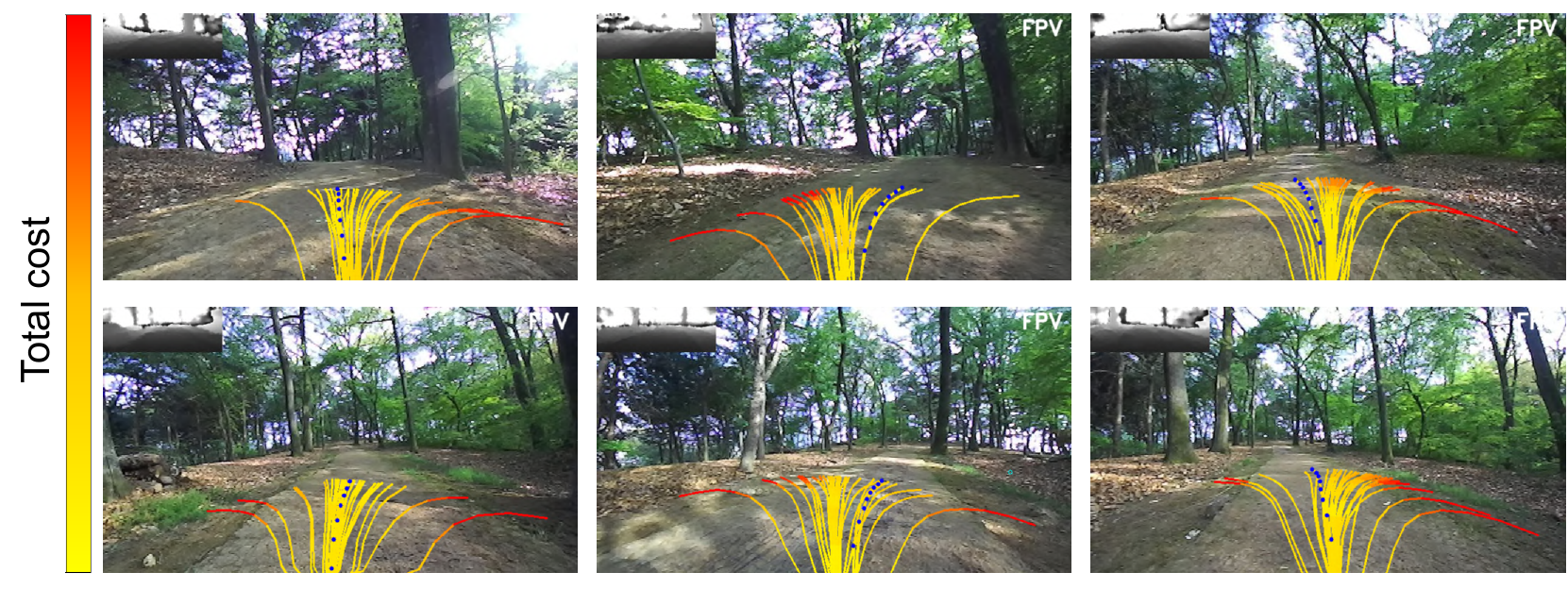}
\caption{Visualized results of deploying Crop-LSTM on the test environment demonstrate its ability to learn without explicit information about the road's edge during training. By learning from high-bumpy driving data with similar visual features, Crop-LSTM accurately predicts the edge's location, resulting in higher costs for that region. Visualized with blue dots, the trajectory with the minimum cost aligns well with human intuition.}
\label{fig_deploy}
\end{figure}

The visual features of the straw blanket and the dirt road are difficult to distinguish, to the extent that even state-of-the-art semantic segmentation models \cite{kirillov2023sam} struggle to detect them accurately.
Despite these difficulties, we observed that the Crop-LSTM predicts the road's edge at a high cost.
Although we did not provide explicit information about the road's edge during training, the Crop-LSTM learned this from the data experiencing drastic drops at the edge of the road.
We show the visualized trajectories, each accompanied by its associated costs, in Fig. \ref{fig_deploy}.

\subsubsection{Success rate}
Next, we demonstrate the effectiveness of our navigation system by testing the success rate for point-goal navigation on unpaved outdoor terrains.
We fixed all elements of the navigation system except the vehicle dynamics model to compare the prediction performance of our Crop-LSTM against BADGR.
By comparing the success rate of the two models, we show the effect of cropping for the navigation task.

\begin{figure*}[!t]
\centering
\includegraphics[width=\textwidth]{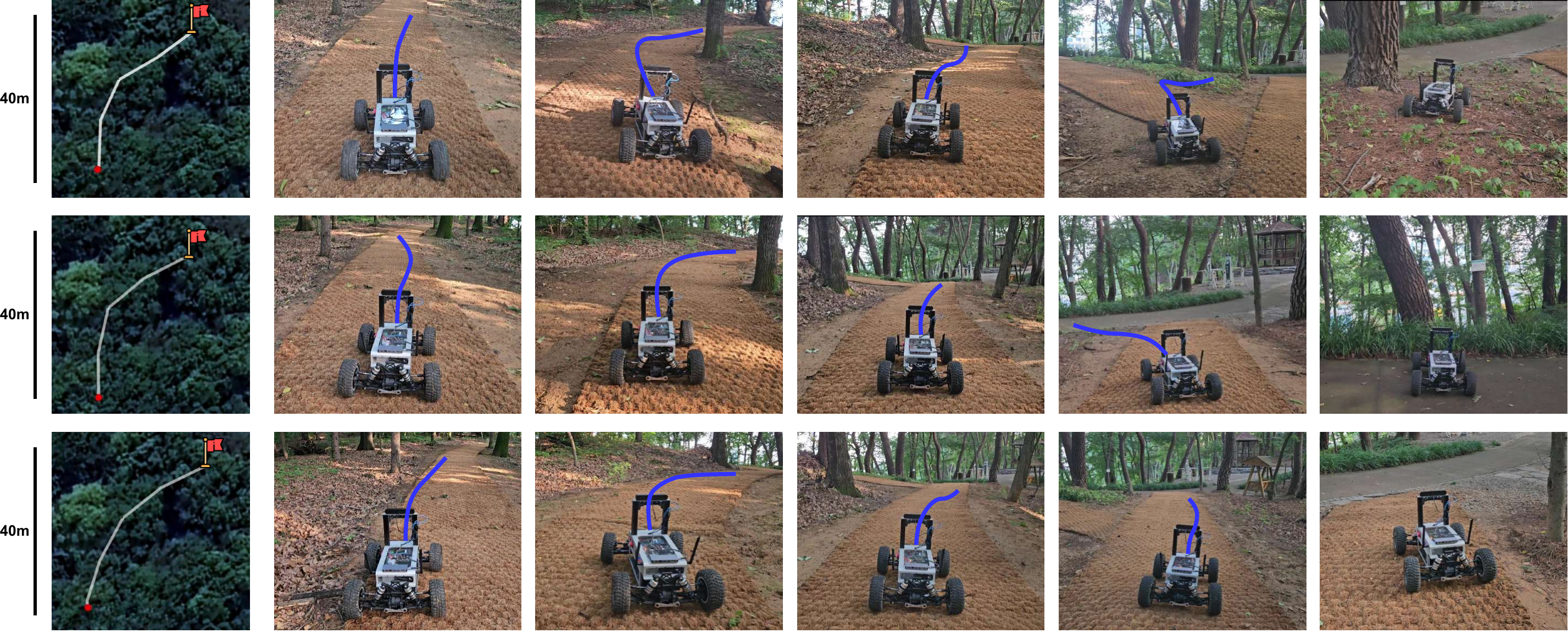}
\caption{We show the navigation results of our proposed navigation system. While driving 45 meters in an unpaved outdoor environment, our navigation system performed path planning and execution. The blue lines show the planned trajectories. We arrived at the goal position without failure.}
\label{fig_navigation}
\end{figure*}

For the point-goal navigation task, the goal position is specified in the robot frame of the starting position. 
We determine the success rate of the point-goal navigation by assessing whether the robot reaches the goal position within a 1m boundary.
In our experiments, we considered two goal distances, 15 meters and 25 meters, while keeping the starting position fixed. 
BADGR failed to drive beyond 25m, preventing a direct comparison with our Crop-LSTM, which successfully navigated the entire 45m course. 
The navigation result for full course driving is shown in Fig. \ref{fig_navigation}.
The success rate was tested at varying distances to the goal position, and the results are presented in Table \ref{table:success_rate}. 

\begin{table}[!ht]
\centering
\caption{Success rate}
\label{table:success_rate}
{\tabcolsep=1.0pt\def\arraystretch{1.0}
\begin{tabularx}{0.65\linewidth}{c *3{>{\Centering}p{42pt}}}
\toprule
  & \multicolumn{2}{c}{Goal distance} \tabularnewline 
  \cmidrule(lr){2-3}
  \textbf{Model} & 15m & 25m \tabularnewline 
\midrule
  BADGR-Modified & 4/10 &  0/5 \tabularnewline
  Crop-LSTM (Ours)  &  \textbf{9/10} & \textbf{4/5} \tabularnewline
\bottomrule
\end{tabularx}}
\end{table}

As shown in Table \ref{table:success_rate}, our method demonstrates superior performance to the baseline, achieving a higher success rate.
Although BADGR has shown impressive results in urban and flat off-road environments, it struggled to adapt to unpaved outdoor environments with narrow straw blankets of width 1.5m. 
During the experiments, BADGR often failed to detect hidden dangers at the road's edge, generating incorrect control commands that get off the straw blanket.
This results in failure since the wheel gets stuck in the fallen leaves.

\begin{figure}[ht]
\centering
\includegraphics[width=\linewidth]{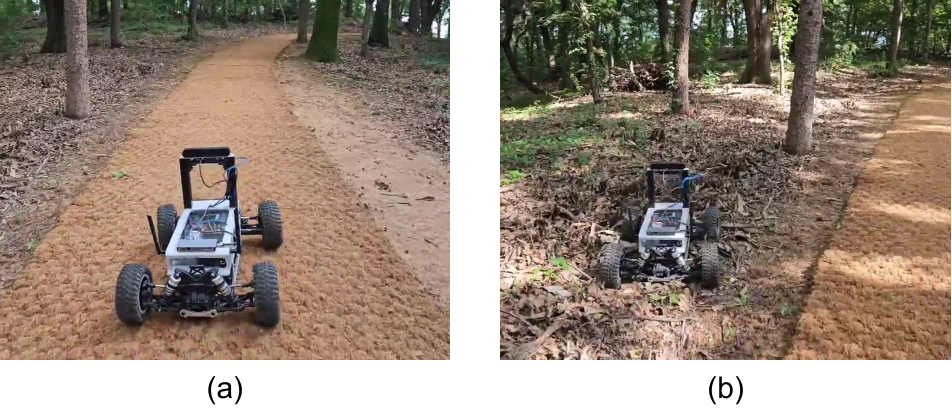}
\caption{(a) While Crop-LSTM avoids getting off to the dirt road, (b) BADGR sometimes fails to correctly predict the hidden dangers.}
\label{fig_crop_success}
\end{figure}

\subsubsection{Bumpiness and Distance driven}
We compare the average bumpiness experienced during driving with the distance driven until termination.
Termination is when \textit{Raicart} either successfully arrives at the goal position, fails by getting off the straw blanket, or collides with a tree. 
The results are presented in Table \ref{table:distance_drived}.

\begin{table}[!ht]
\centering
\caption{Bumpiness and Distance driven}
\label{table:distance_drived}
{\tabcolsep=1.0pt\def\arraystretch{1.0}
\begin{tabularx}{0.95\linewidth}{c *3{>{\Centering}p{80pt}}}
\toprule
  \textbf{Model} & Bumpiness ($\downarrow$) & Distance driven(m) ($\uparrow$) \tabularnewline 
\midrule
  BADGR-Modified & 0.278 & 8.537 \tabularnewline
  Crop-LSTM (Ours) &  \textbf{0.230} & \textbf{20.078} \tabularnewline
\bottomrule
\end{tabularx}}
\end{table}

We observe a reduction of 17.26\% in bumpiness and a 135.18\% increase in the distance driven compared to BADGR. 
The robot experienced less vibration in the z direction and drove further without failure, resulting in steadier and safer navigation.

\begin{figure}[!ht]
\centering
\includegraphics[width=\linewidth]{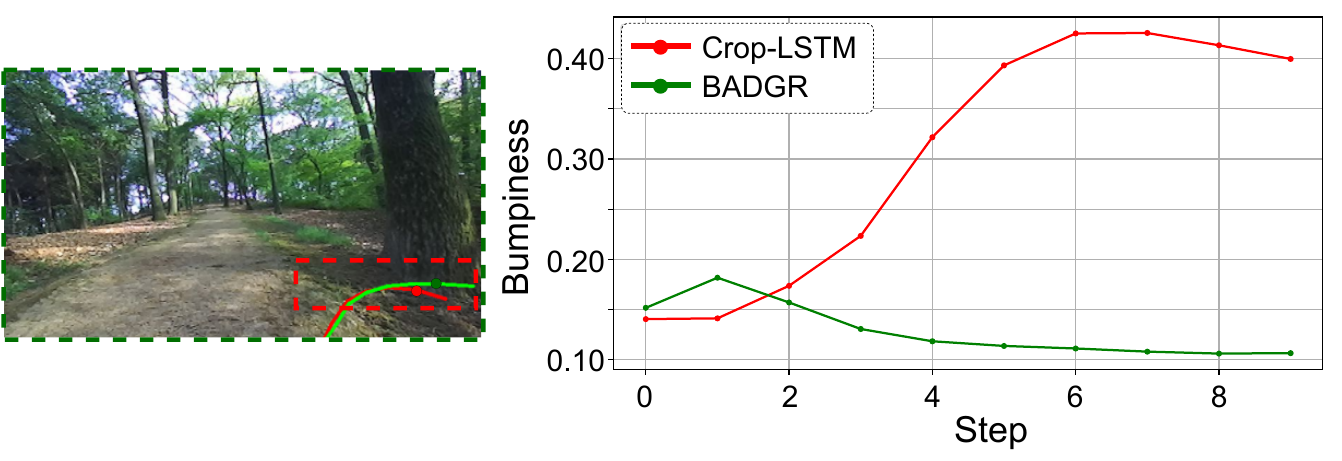}
\caption{Comparison of Crop-LSTM and BADGR in the same test environment. 
It shows similar future trajectories conditioned on the same image and control commands. 
However, predicted bumpiness differs significantly. As the trajectory encounters a tree at step 7, the robot will collide and experience high bumpiness. 
BADGR, utilizing the whole image without attention methods (green dotted), fails to detect the obstacle and predicts lower bumpiness. 
Crop-LSTM's use of cropped features (red dotted) enables accurate vehicle dynamics prediction.}
\label{fig_badgr_crop}
\end{figure}

We discuss the reason why BADGR failed to identify paths that may lead to failure. 
In an environment with a traversable straw blanket and a tree on the right edge, shown in Fig. \ref{fig_badgr_crop}, BADGR did not detect the potential collision risk.
Crop-LSTM could accurately predict upcoming collisions by paying attention to the cropped image patches that have information about the tree.
BADGR focuses more on global information, where the straw blanket comprises a significant portion of the image, leading to a prediction of the low bumpiness of the terrain.
This result shows how our cropping technique improves vehicle dynamics prediction, enabling safe navigation in challenging environments.

\subsection{Obstacle avoidance}
\begin{figure}[ht]
\centering
\includegraphics[width=\linewidth]{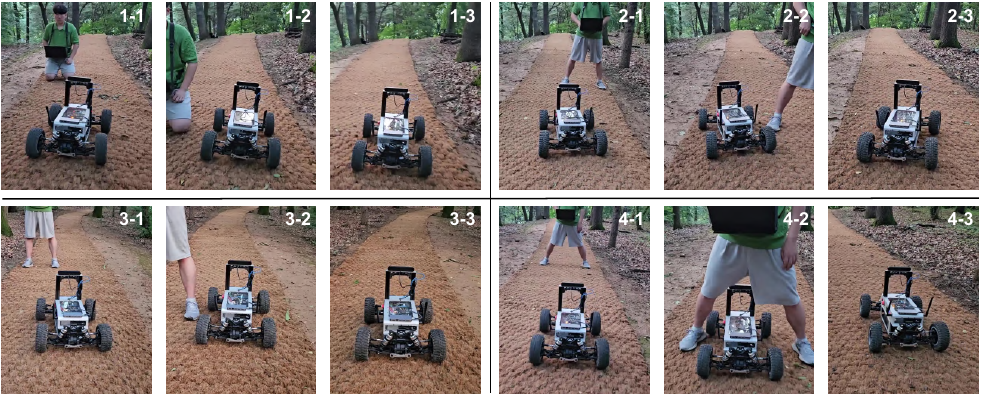}
\caption{Results of obstacle avoidance. When a human is on the side of the straw blanket, the robot navigates to the other side, preventing bumpy events. The robot did not get out of the straw blanket in this evasive maneuver. When both sides are blocked, the robot chooses a path to pass through the legs.}
\label{fig_obstacle}
\end{figure}

We conducted an obstacle avoidance experiment to test our navigation system's performance under unseen disturbances. 
We tested obstacle avoidance for four scenarios where a human is in the environment.
The training data set did not include instances of human presence, so the vehicle dynamics model did not learn about the interactions between the robot and humans.
However, the model predicted potential collisions with nearby objects utilizing the depth camera data. 
We achieved successful obstacle avoidance without modifying any element of the point-goal navigation system, making our approach well-generalized for unseen objects.
Interestingly, the vehicle dynamics model found a narrow traversable path through the legs for a person blocking both edges.
The results of this experiment are shown in Figure \ref{fig_obstacle}.

\section{Conclusion}
In summary, we presented a cropping method for learning an accurate vehicle dynamics model and showed superior performance compared to the previous approach.
We utilized the cropping method and trained a vehicle dynamics model, Crop-LSTM, which pays attention to the region where the robot will be in the future. 
Our navigation system, demonstrated on \textit{Raicart}, successfully navigated challenging outdoor terrain while minimizing bumpiness and preventing failures. 
Furthermore, we showcased the system's capability for obstacle avoidance without requiring any modifications.

Our work is still limited in some aspects.
The T265 tracking camera's odometry data limited \textit{Raicart}'s high-speed travel in the testing environment. 
When the robot drives at high speed, the visual features and the light conditions undergo frequent changes due to the challenging nature of the testing environment, resulting in the instability of the VIO.

We are planning to further extend this work by exploring self-attention-based neural networks. 
Instead of cropping the encoded features on the region where the robot will be in the future, we can utilize self-attention mechanisms to extract the most appropriate features. 
This could improve the prediction accuracy of the vehicle dynamics model, enabling navigation in even more challenging environments.

\bibliographystyle{IEEEtran}
\bibliography{IEEEabrv, ref}

\begin{thebibliography}{10}
\providecommand{\url}[1]{#1}
\csname url@samestyle\endcsname
\providecommand{\newblock}{\relax}
\providecommand{\bibinfo}[2]{#2}
\providecommand{\BIBentrySTDinterwordspacing}{\spaceskip=0pt\relax}
\providecommand{\BIBentryALTinterwordstretchfactor}{4}
\providecommand{\BIBentryALTinterwordspacing}{\spaceskip=\fontdimen2\font plus
\BIBentryALTinterwordstretchfactor\fontdimen3\font minus
  \fontdimen4\font\relax}
\providecommand{\BIBforeignlanguage}[2]{{%
\expandafter\ifx\csname l@#1\endcsname\relax
\typeout{** WARNING: IEEEtran.bst: No hyphenation pattern has been}%
\typeout{** loaded for the language `#1'. Using the pattern for}%
\typeout{** the default language instead.}%
\else
\language=\csname l@#1\endcsname
\fi
#2}}
\providecommand{\BIBdecl}{\relax}
\BIBdecl

\bibitem{fraedrich2019transportation}
E.~Fraedrich, D.~Heinrichs, F.~J. Bahamonde-Birke, and R.~Cyganski,
  ``Autonomous driving, the built environment and policy implications,''
  \emph{Transportation research part A: policy and practice}, vol. 122, pp.
  162--172, 2019.

\bibitem{weisser1999autonomous}
H.~Weisser, P.~Schulenberg, H.~Gollinger, and T.~Michler, ``Autonomous driving
  on vehicle test tracks: overview, implementation and vehicle diagnosis,'' in
  \emph{Proceedings 199 IEEE/IEEJ/JSAI International Conference on Intelligent
  Transportation Systems (Cat. No. 99TH8383)}.\hskip 1em plus 0.5em minus
  0.4em\relax IEEE, 1999, pp. 62--67.

\bibitem{yurtsever2020survey}
E.~Yurtsever, J.~Lambert, A.~Carballo, and K.~Takeda, ``A survey of autonomous
  driving: Common practices and emerging technologies,'' \emph{IEEE access},
  vol.~8, pp. 58\,443--58\,469, 2020.

\bibitem{dolgov2008practical}
D.~Dolgov, S.~Thrun, M.~Montemerlo, and J.~Diebel, ``Practical search
  techniques in path planning for autonomous driving,'' \emph{Ann Arbor}, vol.
  1001, no. 48105, pp. 18--80, 2008.

\bibitem{chu2012local}
K.~Chu, M.~Lee, and M.~Sunwoo, ``Local path planning for off-road autonomous
  driving with avoidance of static obstacles,'' \emph{IEEE transactions on
  intelligent transportation systems}, vol.~13, no.~4, pp. 1599--1616, 2012.

\bibitem{hu2018dynamic}
X.~Hu, L.~Chen, B.~Tang, D.~Cao, and H.~He, ``Dynamic path planning for
  autonomous driving on various roads with avoidance of static and moving
  obstacles,'' \emph{Mechanical systems and signal processing}, vol. 100, pp.
  482--500, 2018.

\bibitem{ostafew2014nonlinear_dynamics_mpc}
C.~J. Ostafew, A.~P. Schoellig, and T.~D. Barfoot, ``Learning-based nonlinear
  model predictive control to improve vision-based mobile robot path-tracking
  in challenging outdoor environments,'' in \emph{2014 IEEE International
  Conference on Robotics and Automation (ICRA)}.\hskip 1em plus 0.5em minus
  0.4em\relax IEEE, 2014, pp. 4029--4036.

\bibitem{williams2016mppi}
G.~Williams, P.~Drews, B.~Goldfain, J.~M. Rehg, and E.~A. Theodorou,
  ``Aggressive driving with model predictive path integral control,'' in
  \emph{2016 IEEE International Conference on Robotics and Automation
  (ICRA)}.\hskip 1em plus 0.5em minus 0.4em\relax IEEE, 2016, pp. 1433--1440.

\bibitem{bojarski2016end-to-end}
M.~Bojarski, D.~Del~Testa, D.~Dworakowski, B.~Firner, B.~Flepp, P.~Goyal, L.~D.
  Jackel, M.~Monfort, U.~Muller, J.~Zhang \emph{et~al.}, ``End to end learning
  for self-driving cars,'' \emph{arXiv preprint arXiv:1604.07316}, 2016.

\bibitem{kebria2019imitation}
P.~M. Kebria, A.~Khosravi, S.~M. Salaken, and S.~Nahavandi, ``Deep imitation
  learning for autonomous vehicles based on convolutional neural networks,''
  \emph{IEEE/CAA Journal of Automatica Sinica}, vol.~7, no.~1, pp. 82--95,
  2019.

\bibitem{kahn2018gcg}
G.~Kahn, A.~Villaflor, B.~Ding, P.~Abbeel, and S.~Levine, ``Self-supervised
  deep reinforcement learning with generalized computation graphs for robot
  navigation,'' in \emph{2018 IEEE international conference on robotics and
  automation (ICRA)}.\hskip 1em plus 0.5em minus 0.4em\relax IEEE, 2018, pp.
  5129--5136.

\bibitem{wellhausen2019should}
L.~Wellhausen, A.~Dosovitskiy, R.~Ranftl, K.~Walas, C.~Cadena, and M.~Hutter,
  ``Where should i walk? predicting terrain properties from images via
  self-supervised learning,'' \emph{IEEE Robotics and Automation Letters},
  vol.~4, no.~2, pp. 1509--1516, 2019.

\bibitem{kahn2021badgr}
G.~Kahn, P.~Abbeel, and S.~Levine, ``Badgr: An autonomous self-supervised
  learning-based navigation system,'' \emph{IEEE Robotics and Automation
  Letters}, vol.~6, no.~2, pp. 1312--1319, 2021.

\bibitem{kendall2019learn_in_a_day}
A.~Kendall, J.~Hawke, D.~Janz, P.~Mazur, D.~Reda, J.-M. Allen, V.-D. Lam,
  A.~Bewley, and A.~Shah, ``Learning to drive in a day,'' in \emph{2019
  International Conference on Robotics and Automation (ICRA)}.\hskip 1em plus
  0.5em minus 0.4em\relax IEEE, 2019, pp. 8248--8254.

\bibitem{shah2021ving}
D.~Shah, B.~Eysenbach, G.~Kahn, N.~Rhinehart, and S.~Levine, ``Ving: Learning
  open-world navigation with visual goals,'' in \emph{2021 IEEE International
  Conference on Robotics and Automation (ICRA)}.\hskip 1em plus 0.5em minus
  0.4em\relax IEEE, 2021, pp. 13\,215--13\,222.

\bibitem{shah2021recon}
D.~Shah, B.~Eysenbach, N.~Rhinehart, and S.~Levine, ``Rapid exploration for
  open-world navigation with latent goal models,'' in \emph{5th Annual
  Conference on Robot Learning}, 2021.

\bibitem{shah2022viking}
D.~Shah and S.~Levine, ``{ViKiNG: Vision-Based Kilometer-Scale Navigation with
  Geographic Hints},'' in \emph{Proceedings of Robotics: Science and
  Systems(RSS)}, 2022.

\bibitem{guan2018crop_diagnose}
Q.~Guan, Y.~Huang, Z.~Zhong, Z.~Zheng, L.~Zheng, and Y.~Yang, ``Diagnose like a
  radiologist: Attention guided convolutional neural network for thorax disease
  classification,'' \emph{arXiv preprint arXiv:1801.09927}, 2018.

\bibitem{dosovitskiy2020vit}
A.~Dosovitskiy, L.~Beyer, A.~Kolesnikov, D.~Weissenborn, X.~Zhai,
  T.~Unterthiner, M.~Dehghani, M.~Minderer, G.~Heigold, S.~Gelly \emph{et~al.},
  ``An image is worth 16x16 words: Transformers for image recognition at
  scale,'' \emph{arXiv preprint arXiv:2010.11929}, 2020.

\bibitem{wang2017residualattention}
F.~Wang, M.~Jiang, C.~Qian, S.~Yang, C.~Li, H.~Zhang, X.~Wang, and X.~Tang,
  ``Residual attention network for image classification,'' in \emph{Proceedings
  of the IEEE conference on computer vision and pattern recognition}, 2017, pp.
  3156--3164.

\bibitem{perreault2020spotnet}
H.~Perreault, G.-A. Bilodeau, N.~Saunier, and M.~H{\'e}ritier, ``Spotnet:
  Self-attention multi-task network for object detection,'' in \emph{2020 17th
  Conference on Computer and Robot Vision (CRV)}.\hskip 1em plus 0.5em minus
  0.4em\relax IEEE, 2020, pp. 230--237.

\bibitem{guan2022ga}
T.~Guan, D.~Kothandaraman, R.~Chandra, A.~J. Sathyamoorthy, K.~Weerakoon, and
  D.~Manocha, ``Ga-nav: Efficient terrain segmentation for robot navigation in
  unstructured outdoor environments,'' \emph{IEEE Robotics and Automation
  Letters}, vol.~7, no.~3, pp. 8138--8145, 2022.

\bibitem{weerakoon2022terp}
K.~Weerakoon, A.~J. Sathyamoorthy, U.~Patel, and D.~Manocha, ``Terp: Reliable
  planning in uneven outdoor environments using deep reinforcement learning,''
  in \emph{2022 International Conference on Robotics and Automation
  (ICRA)}.\hskip 1em plus 0.5em minus 0.4em\relax IEEE, 2022, pp. 9447--9453.

\bibitem{cai2020drift-driving}
P.~Cai, X.~Mei, L.~Tai, Y.~Sun, and M.~Liu, ``High-speed autonomous drifting
  with deep reinforcement learning,'' \emph{IEEE Robotics and Automation
  Letters}, vol.~5, no.~2, pp. 1247--1254, 2020.

\bibitem{romera2017erfnet}
E.~Romera, J.~M. Alvarez, L.~M. Bergasa, and R.~Arroyo, ``Erfnet: Efficient
  residual factorized convnet for real-time semantic segmentation,'' \emph{IEEE
  Transactions on Intelligent Transportation Systems}, vol.~19, no.~1, pp.
  263--272, 2017.

\bibitem{wang2022sfnet}
H.~Wang, Y.~Chen, Y.~Cai, L.~Chen, Y.~Li, M.~A. Sotelo, and Z.~Li, ``Sfnet-n:
  An improved sfnet algorithm for semantic segmentation of low-light autonomous
  driving road scenes,'' \emph{IEEE Transactions on Intelligent Transportation
  Systems}, vol.~23, no.~11, pp. 21\,405--21\,417, 2022.

\bibitem{shah2023lm-nav}
D.~Shah, B.~Osi{\'n}ski, S.~Levine \emph{et~al.}, ``Lm-nav: Robotic navigation
  with large pre-trained models of language, vision, and action,'' in
  \emph{Conference on Robot Learning}.\hskip 1em plus 0.5em minus 0.4em\relax
  PMLR, 2023, pp. 492--504.

\bibitem{goldfain2019autorally}
B.~Goldfain, P.~Drews, C.~You, M.~Barulic, O.~Velev, P.~Tsiotras, and J.~M.
  Rehg, ``Autorally: An open platform for aggressive autonomous driving,''
  \emph{IEEE Control Systems Magazine}, vol.~39, no.~1, pp. 26--55, 2019.

\bibitem{o2020f1tenth}
M.~O'Kelly, H.~Zheng, D.~Karthik, and R.~Mangharam, ``F1tenth: An open-source
  evaluation environment for continuous control and reinforcement learning,''
  \emph{Proceedings of Machine Learning Research}, vol. 123, 2020.

\bibitem{howard2019mobilenetv3}
A.~Howard, M.~Sandler, G.~Chu, L.-C. Chen, B.~Chen, M.~Tan, W.~Wang, Y.~Zhu,
  R.~Pang, V.~Vasudevan \emph{et~al.}, ``Searching for mobilenetv3,'' in
  \emph{Proceedings of the IEEE/CVF international conference on computer
  vision}, 2019, pp. 1314--1324.

\bibitem{kirillov2023sam}
A.~Kirillov, E.~Mintun, N.~Ravi, H.~Mao, C.~Rolland, L.~Gustafson, T.~Xiao,
  S.~Whitehead, A.~C. Berg, W.-Y. Lo \emph{et~al.}, ``Segment anything,''
  \emph{arXiv preprint arXiv:2304.02643}, 2023.

\end{thebibliography}
\end{document}